%% file: main.tex
\documentclass[11pt,a4paper]{article}
\usepackage[hyperref]{naaclhlt2019}

\usepackage{times}
\usepackage{latexsym}

\usepackage{tabularx}
\usepackage{amssymb}
\usepackage{amsmath}
\usepackage{latexsym}
\usepackage{enumitem}
\usepackage{booktabs}
\usepackage{url}
\usepackage{color} 
\usepackage{verbatim}
\usepackage{calc}
\usepackage{arydshln}
\usepackage{mathtools}
\usepackage[draft,textsize=tiny]{todonotes}

\newcommand{\redd}{Reddit$_{13}$}
\usepackage{soul}

\aclfinalcopy 

\title{Short-Term Meaning Shift: A Distributional Exploration}

\author{Marco Del Tredici$^*$ \ \ Raquel Fern\'andez$^*$ \ \  Gemma Boleda$^\dagger$\\
$^*$University of Amsterdam \ \ \ \ $^\dagger$Universitat Pompeu Fabra\\
  {\tt \{m.deltredici|raquel.fernandez\}@uva.nl}\\  
  {\tt gemma.boleda@upf.edu}
}

\date{}

\begin{document}
\maketitle
\begin{abstract}
We present the first exploration of meaning shift over short periods of time in online communities using distributional representations. We create a small annotated dataset and use it to assess the performance of a standard model for meaning shift detection on short-term meaning shift. We find that the model has problems distinguishing meaning shift from referential phenomena, and propose a measure of contextual variability to remedy this.
\end{abstract}

\section{Introduction}
\label{sect:Introduction}
\input{sec_introduction}

\section{Related Work}
\label{sect:Related_Work}
\input{sec_related_work}
\section{Experimental Setup}
\label{sec:setup}
\input{sec_setup}

\section{Types of Meaning Shift}
\label{sec:types}
\input{sec_linguistic_analysis}

\section{Modeling Results and Analysis}
\label{sec:results}

\input{sec_results}
\section{Conclusion}
\label{sect:conc}
\input{sec_conclusion}

\section*{Acknowledgements}
The research carried out by the Amsterdam section of the team was partially funded by the Netherlands Organisation for Scientific Research (NWO) under VIDI grant no.~276-89-008, {\em Asymmetry in Conversation}.
This project has received funding from the European Research Council (ERC) under the European Union’s Horizon 2020 research and innovation programme (grant agreement No 715154), and from the Spanish Ram\'on y Cajal programme (grant RYC-2015-18907). This paper reflects the authors' view only, and the EU is not responsible for any use that may be made of the information it contains.
\begin{flushright}
\includegraphics[width=0.8cm]{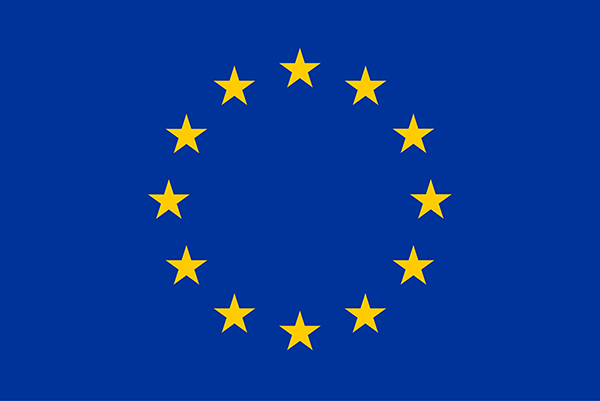}  
\includegraphics[width=0.8cm]{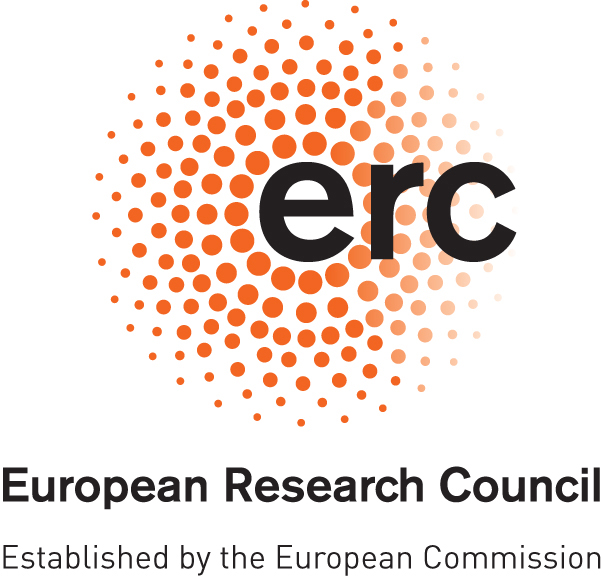} 
\end{flushright}

\bibliography{naaclhlt2019}
\bibliographystyle{acl_natbib}

\appendix

%
%

\section{Further Details on Evaluation Dataset}
\label{sec:further_details_data_model}

For our experiment, we considered content words only, which we identified by using the external list of common words available at \url{https://www.wordfrequency.info/free.asp}.

Three words were discarded from the initial list after analysis of the redditor data: `discord' and `owls' due to the homonymy with proper names not detected during survey's implementation; `tracking' because the chosen examples clearly mislead the judgements of the redditors.

As detailed in Section \ref{subsec:Evaluation dataset}, 26 members of r/LiverpoolFC participated in the survey, and each word received on average 8.8 judgements. We computed inter-annotator agreement as Krippendorff's alpha, and obtained $\alpha$ = 0.58, a relatively low value but common in semantic tasks \cite{artstein2008inter}.

The results of the annotation validate our initial word sampling procedure:

\begin{itemize}
\item  the words that present a significant increase in frequency and were annotated as
meaning shift by us received an average shift annotation of 0.72 ($\pm$ 0.15);
\item the words that present a
significant increase in frequency but that were \emph{not} annotated as
meaning shift by us received an average shift annotation of  0.15 ($\pm$ 0.16);
\item the words that keep a constant frequency between $t_1$ and $t_2$, and we don't consider examples of meaning shift, got 0.07 ($\pm$ 0.12).
\end{itemize}

\end{document}

%% file: sec_introduction.tex
Semantic change has received increasing attention in empirical Computational Linguistics / NLP in the last few years \cite{tang2018state,KutuzovEtal-coling2018}. Almost all studies so far have focused on meaning shift in long periods of time---decades to centuries. However, the genesis of meaning shift and the mechanisms that produce it operate at much shorter time spans, ranging
from the online agreement on words' meaning in dyadic interactions \cite{brennan1996conceptual} to the rapid spread of new meanings in relatively small communities of people in \cite{wenger1998communities,eckert-mcconnellginet1992}.
This paper is, to the best of our knowledge, the first exploration of the latter phenomenon---which we call \textbf{short-term meaning shift}---using distributional representations.

More concretely, we focus on meaning shift arising within a period of 8 years, and explore it on data from an online community of speakers, because there the 
adoption of new meanings happens at a fast pace \cite{Clark96,hasan2009}.
Indeed, short-term shift is usually hard to observe in standard language, such
as the language of books or news, which has been the focus of
long-term shift studies \cite[e.g.,][]{hamilton2016diachronic,kulkarni2015statistically}, since
it takes a long time for a new meaning to be widely accepted in the standard language. 

Our contribution is twofold. First, we create a small dataset of short-term shift for analysis and evaluation, and qualitatively analyze the types of meaning shift we find.%
\footnote{\label{note1}Data and code are available at: 
\url{https://github.com/marcodel13/Short-term-meaning-shift}.} 
This is necessary because, unlike studies of long-term shift, we cannot rely on material previously gathered by linguists or lexicographers.
Second, we test the behavior of a standard distributional model of semantic change when
 applied to short-term shift. 
Our results show that this model successfully detects most shifts in our data, but it overgeneralizes. Specifically, the model gets confused with contextual changes due to speakers in the community often talking about particular people and events, which are frequent on short time spans. 
We propose to use a measure of contextual variability to remedy this and showcase its potential to spot false positives of referential nature like these.
We thus make progress in understanding the nature of semantic shift and towards improving computational models thereof.


%% file: sec_related_work.tex
Distributional models of semantic change are based on the hypothesis
that a change in context of use mirrors a change in meaning.
This in turn stems from the Distributional Hypothesis, that states
that similarity in meaning results in similarity in context of use \cite{harris1954distributional}.
Therefore, all models (including ours) spot semantic shift as a change in the word representation in different time periods.
Among the most widely used techniques are Latent Semantic Analysis \cite{sagi2011tracing,jatowt2014framework}, Topic Modeling \cite{wijaya2011understanding}, classic distributional representations based on co-occurence matrices of target words and context terms \cite{gulordava2011distributional}.
 More recently, researchers have used word embeddings computed using the skip-gram model by \newcite{mikolov2013distributed}. Since embeddings computed in different semantic spaces are not directly comparable, time related representations are usually made comparable either by aligning different semantic spaces through a transformation matrix \cite{kulkarni2015statistically,azarbonyad2017words, hamilton2016diachronic} or by initializing the embeddings at $t_{i+1}$ using those computed at $t_i$ \cite{kim2014temporal,del2016tracing,phillips2017intrinsic,szymanski2017temporal}. We adopt the latter methodology (see Section~\ref{subsec:Model}).

Unlike most previous work, we focus on the language of online communities.
Recent studies of this type of language have investigated the spread of new forms and meanings~\cite{del2017semantic,del2018road,stewart2018making}, 
competing lexical variants \cite{rotabi2017competition}, and the relation between conventions in a community and 
the social standing of its members \cite{danescu2013no}. 
None of these works has analyzed the ability of a distributional model to capture these phenomena, 
which is what we do in this paper for short-term meaning shift. 
\newcite{kulkarni2015statistically} consider meaning shift in short time periods on Twitter data, but without providing an analysis of the observed shift nor systematically assessing the performance of the model, as we do here.

Evaluation of semantic shift is difficult, due to the lack of
annotated datasets \cite{frermann2016bayesian}. For this reason, even for long-term shift, evaluation is usually performed by manually
inspecting the $n$ words whose representation changes the most
according to the model under investigation~\cite{hamilton2016diachronic,kim2014temporal}.
Our dataset allows for a more systematic evaluation and analysis, and
enables comparison in future studies.

%% file: sec_setup.tex
\subsection{Data}
\label{subsec:Data}
We exploit user-generated language from an online forum of football fans,
namely, the r/LiverpoolFC subreddit, one of the many communities hosted
by the Reddit platform.\footnote{\url{https://www.reddit.com}. We downloaded Reddit data using the Python package Praw: \url{https://pypi.python.org/pypi/praw/}.}
\newcite{del2018road} showed that this subreddit presents many characteristics that favour the creation and spread of linguistic innovations, such as a topic that reflects a strong external interest and high density of the connections among its users. This makes it a good candidate for our investigation.
We focus on a short period of eight years, between 2011 and 2017. 
In order to enable a clearer observation of short-term meaning shift, we define two
non-consecutive time bins: the first one ($t_1$) contains data from
2011--2013 and the second one ($t_2$) from 2017.\footnote{These choices
  ensure that the samples in these two time bins are approximately of the same size -- see Table~\ref{tab:data}. The
  r/LiverpoolFC subreddit exists since 2009, but very little content
  was produced in 2009--2010.}
 We also use a large sample of community-independent language for the
initialization of the word vectors, namely, a random crawl from Reddit
in 2013.
Table~\ref{tab:data} shows the size of each sample.

\subsection{Model}
\label{subsec:Model}
We adopt the model proposed by \newcite{kim2014temporal}, a representative method for computing diachronic meaning shift.\footnote{The model was implemented using the Python package Gensim: \url{https://pypi.python.org/pypi/gensim/}.} While other methods might be equally suitable (see Section \ref{sect:Related_Work}), we expect our results not to be method-specific, because they concern general properties of short-term shift, as we show in Sections~\ref{sec:types} and~\ref{sec:results}.
In the model by \newcite{kim2014temporal}, word
embeddings for the first time bin $t_1$ are initialized randomly; then,
given a sequence of time-related samples, embeddings for $t_i$
are initialized using the embeddings of $t_{i-1}$ and further
updated. 
If at $t_i$ the word is used in the same contexts as in $t_{i-1}$, its embedding will only be marginally updated, whereas a major change in the context of use will lead to a stronger update of the embedding. The model makes embeddings across time bins directly comparable.

We implement the following steps:
First, we create  randomly initialized word embeddings with the large sample \redd\ to obtain 
meaning representations that are community-independent.
We then use these embeddings
to initialize those in LiverpoolFC$_{13}$, update the vectors on this
sample, and thus obtain embeddings for time $t_1$. This step
adapts the general embeddings 
to the LiverpoolFC community. Finally, we
initialize the word embeddings for LiverpoolFC$_{17}$ with those of
$t_1$, train on this sample, and get embeddings for $t_2$.


The vocabulary is defined as the intersection of the
vocabularies of the three samples (\redd, LiverpoolFC$_{13}$,
LiverpoolFC$_{17}$), and includes 157k words.
For \redd, we include only words that occur at least 20 times in the
sample, so as to ensure meaningful representations for each word,
while for the other two samples we do not use any frequency
threshold: Since the embeddings used for the initialization of
LiverpoolFC$_{13}$ encode community-independent meanings, if a word doesn't occur in
LiverpoolFC$_{13}$ its representation will simply be as in \redd,
which reflects the idea that if a word is not used in a community, then its meaning is not altered within
that community. 
We train with standard skip-gram parameters \cite{levy2015improving}: window 5, learning rate 0.01, embedding dimension 200, hierarchical softmax.

\begin{table}[t!]
\centering
\begin{tabular}{lccc}
\bf sample & \bf time bin & \bf million tokens \\
 \hline
\redd &  2013 & {\raise.17ex\hbox{$\scriptstyle\sim$}}900 \\
LiverpoolFC$_{13}$ & 2011--13 & ~ 8.5\\
LiverpoolFC$_{17}$ & 2017 & 11.9\\ \hline
\end{tabular}
\caption{Time bin and size of the datasets.}
\label{tab:data}
\end{table}


\subsection{Evaluation dataset}
\label{subsec:Evaluation dataset}

Our dataset consists of 97 words from the r/LiverpoolFC subreddit with
annotations by members of the subreddit ---that is, community members
with domain knowledge (needed for this task) but no linguistic
background.

To ensure that we would get enough cases of semantic shift to enable a
meaningful analysis, we started out from content words that increase
their relative frequency between $t_1$ and $t_2$.\footnote{Frequency
  increase has been shown to positively correlate with meaning change
  \cite{wijaya2011understanding,kulkarni2015statistically}; although
  it is not a necessary condition, it is a reasonable starting point,
  as a random selection of words would contain very few positive
  examples. Our dataset is thus biased towards precision over recall.}
A threshold of 2 standard deviations above the mean yielded
$\sim$200 words.~The first author manually identified 34 semantic
shift candidates among these words by analyzing their contexts of use
in the r/LiverpoolFC data.  Semantic shift is defined here as a change
in the ontological type that a word denotes, which takes place when the word starts to be used to denote an entity which is different from the one originally denoted and the new use spreads among the members of a community (see examples in Sec.~\ref{sec:types}).
We added two types of confounders: 33 words
with a significant frequency increase but not marked as meaning
shift candidates, and 33 words with constant frequency between $t_1$
and $t_2$, included as a sanity check. All words have
  absolute frequency in range [50--500].

The participants were shown the 100 words (in randomized order)
together with randomly chosen contexts of usage from each time period
($\mu$=4.7 contexts per word) and, for simplicity, were asked to make
a binary decision about whether there was a change in meaning. In order to have the redditors familiarize with the concept of meaning change, we first provide them with an intuitive, non-technical definition, and then a set of cases that exemplify it. The instructions to participants can be found in the project's GitHub repository (see footnote \ref{note1}).

Semantic shift is arguably a graded notion. In line with a suggestion by \newcite{KutuzovEtal-coling2018} to account for this fact, we aggregate the annotations into a graded \emph{semantic shift index},
ranging from 0 (no shift) to 1 (shift) depending on how many subjects
spotted semantic change. The shift index is exclusively based
  on the judgments of the redditors, and does not consider the
  preliminary candidate selection done by us.  Overall, 26 members
of r/LiverpoolFC participated in the survey, and each word received on
average 8.8 judgements. 
  Further details about the dataset are in Appendix \ref{sec:further_details_data_model}.


%% file: sec_linguistic_analysis.tex
We identify three main types of shift in our data via qualitative analysis of examples with a shift index $> 0.5$: metonymy, metaphor, and meme.  

\paragraph{Metonymy.}
In metonymic shifts, a highly salient characteristic of an entity is used to refer to it. Among these cases are, for example, {\em `highlighter'}, which in $t_2$ occurs in sentences like \textit{`we are playing with the highlighter today'}, or \textit{`what's up with the hate for this kit? This is great, ten times better than the highlighter'},
used to talk about a kit in a colour similar to that of a highlighter pen; or {\em `lean'}, in \textit{`I hope a lean comes soon!'}, \textit{`Somebody with speed\dots make a signing\dots Cuz I need a lean'},
used to talk about hiring players due to new hires typically leaning on a Liverpool symbol when posing for a photo right after signing for the club. Particularly illustrative is the `F5' example shown in
Table~\ref{table:f5}. While `F5' is initially used with its common usage of shortcut for refreshing a page (1), it then starts to denote the act of refreshing in order to get the latest news about the possible transfer of a new player to LiverpoolFC (2). This use catches on and many redditors use it to express their tension while waiting for good news (3-5),\footnote{Here the semantic change is accompanied by a change in the part of speech, and `F5' becomes a denominal verb.}
though not all members are aware of the new meaning of the word (6). When the transfer is almost done, someone leaves the {\em `F5 squad'} (7), and after a while, another member recalls the period in which the word was used (8).

\begin{table}[t]\centering
    \begin{tabular}{@{}c@{\ \ }p{5.5cm}r@{}}
        \hline
        (1) & \em Damn, after losing the F5 key on my keyboard [...] & 16 Jun\\\hline
        (2) & \em [he is] so  close, F5 tapping is so intense right now & 18 Jun\\\hline
        (3) & \em Don't think about it too much, man. Just F5 & 1 Jul\\\hline
        (4) & \em Literally 4am I slept and just woke up and thought it was f5 time & 3 Jul\\\hline
        (5) & \em this was a happy f5 & 3 Jul\\\hline
        (6) & \em what is an F5? & 3 Jul \\\hline
        (7) & \em I'm leaving the f5 squad for now & 5 Jul\\\hline
        (8) & \em I made this during the f5 madness & 6 Sep\\\hline      
    \end{tabular}
    \caption{Examples of use of `F5' with time stamps, which illustrate the speed of the meaning shift process. All the examples are from LiverpoolFC$_{17}$.}
     \label{table:f5}
\end{table}

\paragraph{Metaphor.}
Metaphorical shifts lead to a broadening of the original meaning of a
word through analogy.
For example, in $t_2$ {\em `shovel'} occurs in sentences such as \textit{`welcome aboard, here is your shovel'} or 
\textit{`you boys know how to shovel coal'}: the team is seen as a train that is running through the season, and every supporter is asked to figuratively contribute by shoving coal into the train boiler. 

\paragraph{Meme.}
Finally, memes are another prominent source of meaning shift. 
In this case, fans use a word to make jokes and be sarcastic, and the new usage quickly spreads within the community. 
As an example, while Liverpool was about to sign a new player named Van Dijk, redditors started to play with the homography of the first part of the surname with the common noun `van', its plural form `vans', and the shoes brand `Vans': `\textit{Rumour has it Van Djik was wearing these vans in the van}' or `\textit{How many vans could Van Dijk vear if Van Dijk could vear vans}'. 
Jokes of this kind are positively received in the community (`\textit{Hahah I love it. Anything with vans is instant karma!}') and quickly become frequent in it.


%% file: sec_results.tex
\begin{figure}[t]\centering
\includegraphics[width=\columnwidth]{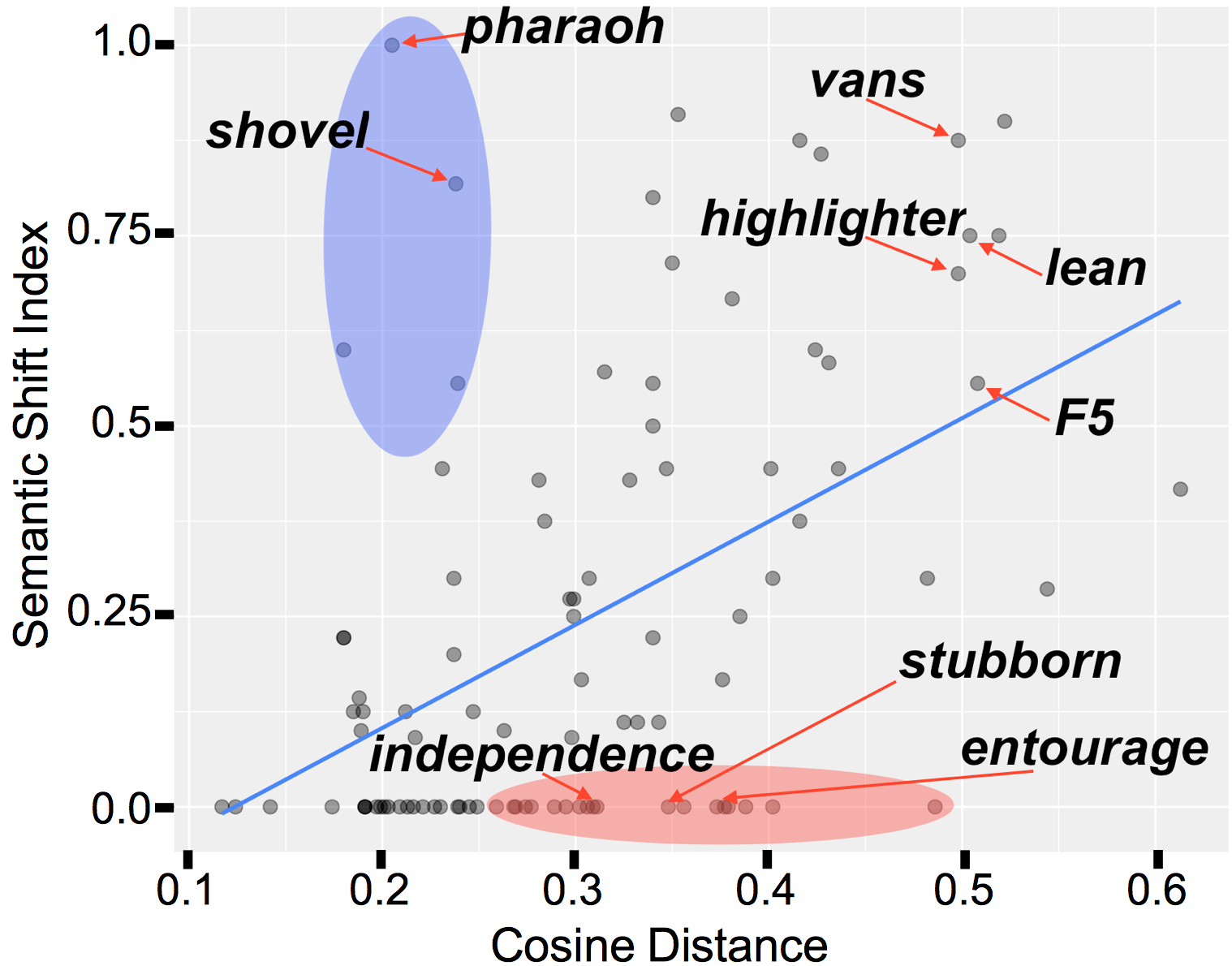}
\caption{Semantic shift index vs.~cosine distance in the evaluation dataset (Pearson's $r$ = 0.49, $p< 0.001$). 
Red horizontal ellipsis:~false positives; blue vertical ellipsis:~false negatives.
\label{fig:shift-cosine}}
\end{figure}

The positive correlation between cosine distance and semantic shift index (Pearson's $r$= 0.49, $p<0.001$, see Figure \ref{fig:shift-cosine}) confirms the hypothesis that meaning shift is mirrored by a change in context of use. However, we also find systematic deviations. 

\subsection{False negatives}
\label{subsec:False negatives}

A small, but consistent group is that of 
words that undergo semantic shift but are not captured by the model
(blue vertical ellipsis Figure~\ref{fig:shift-cosine}; shift index\textgreater 0.5, cosine distance\textless 0.25).
These are all metaphorical shifts; in particular, cases of extended
metaphor \cite{werth1994extended}, where the metaphor is 
developed throughout the whole text.
For instance, besides the {\em `shovel'} example mentioned in Section~\ref{sec:types}, we find {\em `pharaoh'}, the nickname of an Egyptian player who
joined Liverpool in 2017, used in
sentences like \textit{`approved by our new Pharaoh Tutankhamun'}, or \textit{`our dear Egyptian Pharaoh, let's hope he becomes a God'}.
Despite the metaphoric usage, the local context of these words is similar to the literal one, and so the model does not spot the meaning shift. We expect this to happen in long-term shift models, too, but we are not aware of results confirming this.

\subsection{False positives}
\label{subsec:False positives}
A larger group of problematic cases is that of 
words that do \textit{not} undergo semantic shift despite showing
relatively large differences in context between $t_1$ and $t_2$ (red horizontal ellipsis in
Figure~\ref{fig:shift-cosine}; shift index=0, cosine distance\textgreater	
0.25). Manual inspection reveals that most of these ``errors'' 
are due to a referential effect: words are used
almost exclusively to refer to a specific person or event in $t_2$, and
so the context of use is different from the contexts in $t_1$.
For instance, {\em `stubborn'} is
almost always used to talk about a coach who was not
there in 2013 but only in 2017; 
{\em `entourage'}, for the entourage of a particular star of the team; {\em `independence'} for the
political events in Catalonia (Spain). 
In all these cases, the meaning of the word stays the same, 
despite the change in context. In line with the Distributional
Hypothesis, the model spots the context change, but it is not
sensitive to its nature. We expect long-term shift to not be as
susceptible to referential effects 
like these because
embeddings are aggregated over a larger and more varied number of
occurrences.

\begin{figure}[h!]\centering
\includegraphics[width=\columnwidth]{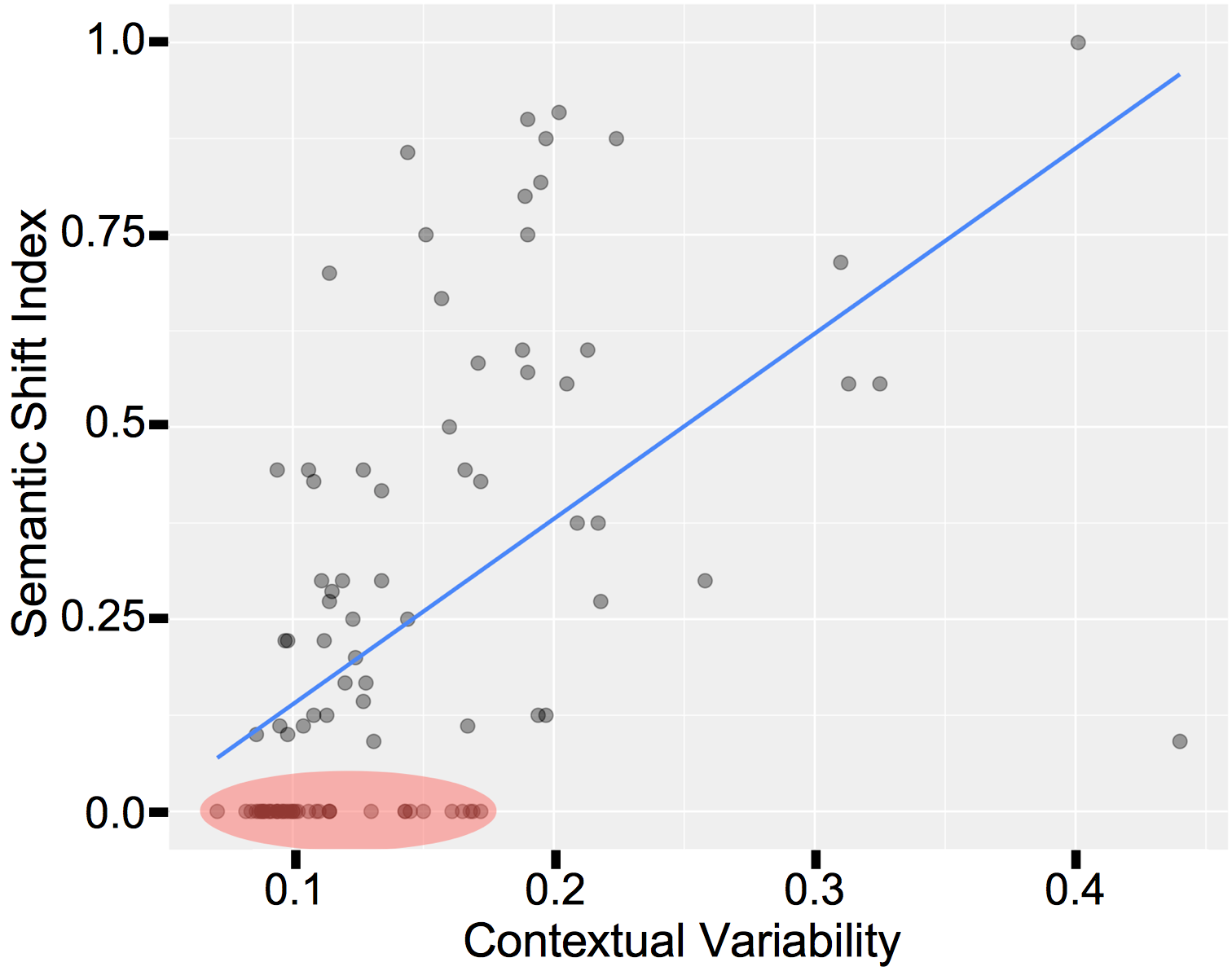}
\caption{Semantic shift index vs.~context variability. Red horizontal ellipsis: referential cases which are assigned high cosine distance values by the model (false positives).\label{fig:shift-variability}}
\end{figure}

We expect that in referential cases the contexts of use will 
be \textit{narrower} than for words with actual
semantic shift, as they are specific to one person or
event. Hence, a measure of \textit{contextual variability}
should help spot false positives.  To test this hypothesis, we define
contextual variability as follows: For a target word, we create a
vector for each of its contexts (5 words on both sides of the target) in $t_2$ by averaging the embeddings of
the words occurring in it, and define variability as the average
pairwise cosine distance between context vectors.%
\footnote{There are alternative ways of measuring contextual variability, but we expect them to yield the same picture.
For instance, we experimented with a different window size and obtained the same pattern.}
We find that
contextual variability is indeed significantly correlated with
semantic shift in our dataset (Pearson's $r\!=\!0.55$, $p\!<\!0.001$),
while it is independent from cosine distance (Pearson's $r$= 0.18,
$p> 0.05$). These two aspects are thus complementary. While both shift
words and referential cases change context of use in $t_2$, context
variability captures the fact that only in referential cases words
occur in a restricted set of contexts. 
Figure~\ref{fig:shift-variability} shows this effect visually.  This
result can inform future 
models of short-term meaning
shift.

%% file: sec_conclusion.tex
The goal of this initial study was to bring to the attention of the NLP community {\bf short-term meaning shift}, an under-studied problem in the field. 
Our hope is that it will spark further research into a phenomenon which, besides being of theoretical interest, has potential practical implications for NLP downstream tasks concerned with user-generated language, as modeling how word meanings rapidly change in communities would allow a better understanding of what their members say.
Future research should experiment with other datasets (reddits from other domains, other online communities) and also alternative models that address the challenges described here.
